\documentclass[lettersize,journal]{IEEEtran}
\usepackage[caption=false,font=normalsize,labelfont=sf,textfont=sf]{subfig}
\usepackage{textcomp}
\usepackage{stfloats}
\usepackage{url}
\usepackage{verbatim}
\usepackage{amsthm,amsmath, amssymb, mathtools, amsfonts}
\usepackage{ulem}
\usepackage{bm}
\usepackage{microtype}
\usepackage{graphicx}
\usepackage{comment}
\usepackage{makecell}
\usepackage{booktabs} 
\usepackage{stackrel}
\usepackage{comment}
\usepackage{hyperref}
\usepackage{xr}
\usepackage{xcolor}
\definecolor{crimson}{rgb}{0.86, 0.08, 0.24}
\definecolor{hanblue}{rgb}{0.27, 0.42, 0.81}
\definecolor{amethyst}{rgb}{0.6, 0.4, 0.8}

\newcommand{\av}[1]{\left\langle {#1} \right\rangle }

\def\BibTeX{{\rm B\kern-.05em{\sc i\kern-.025em b}\kern-.08em
    T\kern-.1667em\lower.7ex\hbox{E}\kern-.125emX}}
\usepackage{balance}

\begin{document}
\title{Fast and Functional Structured Data Generators Rooted in Out-of-Equilibrium Physics}

\author{\IEEEauthorblockN{Alessandra Carbone\IEEEauthorrefmark{1},
Aurélien Decelle\IEEEauthorrefmark{2}\IEEEauthorrefmark{3}, Lorenzo Rosset\IEEEauthorrefmark{2}\IEEEauthorrefmark{3} and
Beatriz Seoane\IEEEauthorrefmark{2}\IEEEauthorrefmark{3}}

\IEEEauthorrefmark{1}\IEEEauthorblockA{Sorbonne Université, CNRS, IBPS, Laboratoire de Biologie Computationnelle et Quantitative - UMR 7238, 75005 Paris, France.}

\IEEEauthorrefmark{2}\IEEEauthorblockA{Departamento de Física Teórica, Universidad Complutense de Madrid, 28040
Madrid, Spain.}

\IEEEauthorrefmark{3}\IEEEauthorblockA{Université Paris-Saclay, CNRS, INRIA Tau team, LISN, 91190 Gif-sur-Yvette,
France.}\\
E-mail: lorenzo.rosset@phys.ens.fr}

\markboth{Journal of \LaTeX\ Class Files,~Vol.~18, No.~9, September~2020}%
{How to Use the IEEEtran \LaTeX \ Templates}

\maketitle

\begin{abstract}
In this study, we address the challenge of using energy-based models to produce high-quality, label-specific data in complex structured datasets, such as population genetics, RNA or protein sequences data. Traditional training methods encounter difficulties due to inefficient Markov chain Monte Carlo mixing, which affects the diversity of synthetic data and increases generation times. To address these issues, we use a novel training algorithm that exploits non-equilibrium effects. This approach, applied to the Restricted Boltzmann Machine, improves the model's ability to correctly classify samples and generate high-quality synthetic data in only a few sampling steps. The effectiveness of this method is demonstrated by its successful application to five different types of data: handwritten digits, mutations of human genomes classified by continental origin, functionally characterized sequences of an enzyme protein family, homologous RNA sequences from specific taxonomies and real classical piano pieces classified by their composer.
\end{abstract}


\section{Introduction}
In recent years, many branches of science have made rapid progress through the development and implementation of generative machine learning models. Among the most notable examples are large language models in natural language processing such as ChatGPT~\cite{openai2023chatgpt} or music generation~\cite{copet2024simple}, or their application in fields such as medical diagnostics~\cite{natarajan2023medpalm} and protein structure prediction~\cite{jumper2021alphafold}. In addition, neural networks are revolutionizing the simulations of quantum mechanics~\cite{carleo2017solving} and materials science~\cite{li2022deep} by representing complex many-body states with unprecedented accuracy.

In biology, significant progress has been made recently in designing artificial protein sequences with desired properties using deep learning networks \cite{alley2019unified, shin2021protein, anishchenko2021novo, moffat2022design}. The problem with these approaches is that they require a lot of data to train and are unable to generate sequences at the resolution level of subfamily-specific features due to the limited amount of data. Automatic classification of protein sequences according to their biological function based on a few examples is also a complex task because the function cannot be inferred directly from phylogeny~\cite{pazos2006phylogeny}, and other sequence spaces have been
defined for this end~\cite{vicedomini2022multiple,decelle2023unsupervised}.
In this work, we address the problem of classifying and generating sequences with a particular labeling prescription in a more general context using energy-based models (EBMs).

EBMs~\cite{ackley1985learning,smolensky1986information,lecun2006tutorial,xie2016theory} are powerful generative models that encode a complex dataset distribution into the Boltzmann distribution of a given energy function. Their simplest versions, the Boltzmann~\cite{ackley1985learning} and the Restricted Boltzmann~\cite{smolensky1986information} machines, have recently got renewed attention in the scientific world, not only because they can generate high-quality synthetic samples in datasets for which convolutional layers offer no appreciable advantage~\cite{cocco2018inverse,yelmen2021creating,yelmen2023deep}, but also because they offer appealing modelling and interpretation capabilities while requiring relatively small training sets. Indeed, the trained model can be understood and studied as a physical interaction system to model many-body distributions~\cite{carleo2017solving,melko2019restricted}, infer physical interactions~\cite{weigt2009identification,morcos2011direct}, extract patterns~\cite{tubiana2019learning}, or cluster~\cite{decelle2023unsupervised}. The process of feature encoding can also be analytically rationalized to some extent~\cite{decelle2021restricted,decelle2017spectral}.
However, EBMs pose a major difficulty in training because the goodness of the trained models depends entirely on the quality of convergence to equilibrium of the Markov Chain Monte Carlo (MCMC) sampling used to estimate the log-likelihood gradient during training~\cite{decelle2021equilibrium,nijkamp2022mcmc}. These concerns are particularly critical when dealing with highly structured datasets, as sampling multimodal distributions is very costly. This is because the mixing times increase rapidly during training, which is dominated by barriers between metastable states. Non-ergodic MCMC sampling often leads to models that overrepresent certain modes at the equilibrium distribution level~\cite{Nijkamp_Hill_Han_Zhu_Wu_2020,decelle2021equilibrium}.
This is also true when subsequent MCMC processes are initialized with the chain states used to compute the previous gradient update, the so-called persistent contrastive divergence (PCD) recipe~\cite{tieleman2008training}. Moreover, even perfectly trained models can be poor generators as they cannot reproduce the full diversity encoded in the probability measure because the chains cannot mix in a reasonable amount of time.

Recent works~\cite{nijkamp2019learning,decelle2021equilibrium,agoritsas2023explaining} have shown that if the goal is to generate new samples, it is easier to train EBMs using a non-convergent approach, rather than to adjust the parameters of the Boltzmann distribution in such a way that equilibrium generated samples match the empirical distribution of the dataset.
This means that EBMs can be trained to work as diffusion models~\cite{sohl2015deep}, i.e. as fast and accurate generators that perform a set of decoding tasks that can be imposed on the model during training. For structured datasets, this strategy offers two obvious advantages: The generated samples better reflect the diversity of the dataset, and the number of MCMC steps required to generate high-quality samples can be very short. In addition, training EBMs out-of-equilibrium is not only faster than the standard procedure, but also more stable and easier to control~\cite{decelle2021equilibrium}. Since this training strategy has largely been tested on images, it has yet to be explored with highly structured datasets where thermalization is prohibitively expensive.

\begin{figure*}[!ht]
    \centering
    \includegraphics[width=0.9\textwidth]{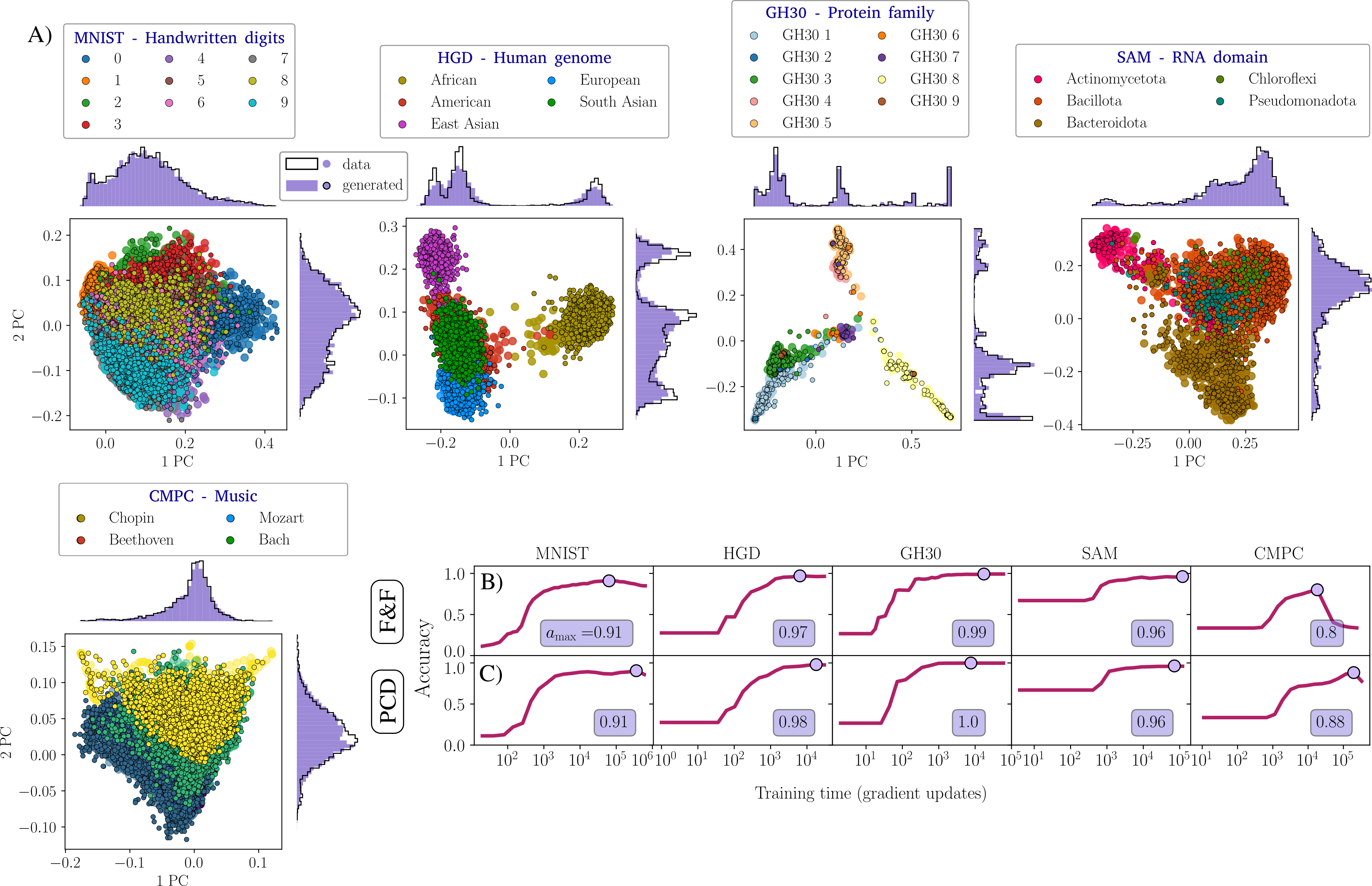}
    \caption{A) Our F\&F-10 model performs accurate conditional generation on a variety of datasets, as shown from left to right:(MNIST) handwritten digits classified by number, (HGD) mutations in the human genome classified by continental origin, (GH30) sequences from a homologous enzyme protein family characterized by different biological functions, (SAM) a homologous family of RNA sequences classified by taxonomy and finally (CMPC), a collection of piano pieces by various composers. The generated samples are obtained by sampling the model equilibrium distribution for only 10 MCMC steps from a random initialization. The real and fake data are projected along the two principal components of the PCA of the dataset. Large dots indicate real data, while smaller contoured dots represent generated samples. Each color corresponds to a particular label. The synthetic dataset mirrors the structure of the real dataset, ensuring that each category has exactly the same number of entries. The histograms in the outer panels illustrate the distributions of the dataset (black outline) and the generated samples (violet-shaded area), projected along each of the main directions shown in the central scatter plot. Comparison of the accuracy's of B): F\&F and C): PCD RBM in predicting the labels of samples in the test set as a function of the training time. The inference is done by starting from an initial random label and then performing $10^3$ MCMC steps. The purple box in the corner of the insets indicates the maximum accuracy achieved ($a_{\mathrm{max}}$), corresponding to the big purple dot.}
    \label{fig:PCAs hist Rdm}
\end{figure*}

In this paper, we show that Restricted Boltzmann Machines (RBMs) can be simultaneously trained to perform two different tasks after a few MCMC sweeps (just 10 in our experiments). First, they are able to generate samples conditioned on a particular label when initialized with random conditions. The samples generated by the model satisfy the individual label statistics with high accuracy and cover the entire data space in the correct proportion (Fig.~\ref{fig:PCAs hist Rdm}-A). Second, they can accurately predict the label associated with a given sample (Fig.\ref{fig:PCAs hist Rdm}-B). We validate our method on five different datasets: images of handwritten digits (MNIST), primarily to illustrate the method, and four highly structured datasets of different types: one listing human DNA mutations in individuals, two others containing sequences of homologous protein and RNA families respectively, and a final one containing classical piano pieces integrating time correlations. All these datasets are introduced later in the manuscript, and more details can be found in the supplementary information (SI).
For the latter datasets, generating high-quality and feature-dependent data is usually challenging, if not impossible, in a reasonable amount of time.

The structure of this paper is as follows: we begin by introducing our EBM and the out-of-equilibrium training protocol. We then discuss our results in detail, coupled with an analysis of the tests performed to assess the quality of the generated samples. The paper concludes with a summary of our results and conclusions.

\section{Restricted Boltzmann Machine}\label{Restricted Boltzmann Machine}
Although RBMs have been around for a long time, they are largely used to describe aligned DNA, RNA or protein sequence datasets~\cite{tubiana2019learning,bravi2021rbm,yelmen2021creating,yelmen2023deep,fernandez2023designing}. There are two reasons for this. First, convolutional layers are unlikely to provide much advantage in this case, and most importantly, they do not require many training examples to provide reliable results. The latter is especially important when dealing with semi-supervised tasks, since the number of manually curated entries is usually very small compared to the number and diversity of sequences available in public databases. We will devote all our work to this type of tasks and machines.

\subsection{Definition of the model}
The RBM is a Markov random field with pairwise interactions defined on a bipartite graph of two noninteracting layers of variables: the visible variables $\pmb{v} = \{v_i\}_{i=1, \dots, N_v}$ represent the data, while the hidden variables $\pmb{h} = \{h_\mu\}_{\mu = 1, \dots, N_h}$ form a latent representation o
that models the effective interactions between the visible variables. The joint probability distribution of the visible and hidden variables is given by the Boltzmann distribution
\begin{equation}
    p_{\pmb{\theta}}(\pmb{v}, \pmb{h}) =\textstyle\frac{1}{Z_{\pmb{\theta}}} e^{- E_{\pmb{\theta}}(\pmb{v}, \pmb{h})}\text{  with  } Z_{\pmb{\theta}} =\textstyle \sum_{\pmb{v}, \pmb{h}} e^{- E_{\pmb{\theta}}(\pmb{v}, \pmb{h})}.
    \label{eq:Boltzmann distribution}
\end{equation}
In the previous expressions, the normalization factor $Z_{\pmb{\theta}}$ is called the \textit{partition function}, $\pmb{\theta}$ refers to the parameters of the model and $E$ is the energy function or \textit{Hamiltonian}.
In the simplest case, both the visible and the hidden units are binary variables, $v_i, h_\mu \in \{0, 1\}$, but we will also consider categorical (namely Potts) variables for $v_i$ in the case of the homologous protein and RNA sequence datasets, see e.g.~\cite{tubiana2019learning,decelle2023unsupervised} for a Potts version of the model. For the semi-supervised setting, we introduce an additional categorical variable in the visible layer, $\ell \in \{1, \dots, N_\ell\}$, that represents the label associated with the data point. That is, we follow the same scheme as in Ref.~\cite{larochelle2012learning}, but use a categorical variable for the label instead. 
We note that a very different procedure for conditional generation in RBMs have recently been proposed~\cite{fernandez2023disentangling}. 
The associated Hamiltonian is
\begin{align}\label{eq:hamiltonian}
     E_{\pmb{\theta}}(\pmb{v}, \pmb{h}, \ell) \!=\! \textstyle-&\!\sum_{i} a_i v_i \! -\! \sum_\mu b_\mu h_{\mu} \!-\! \sum_{i \mu}v_i w_{i \mu } h_\mu \nonumber\\
    &\textstyle - \sum_m c_m \delta_{\ell, m} \!-\! \sum_{m \mu} \delta_{\ell, m} d_{m \mu} h_\mu,
\end{align}
where $\delta_{\ell, m}$ is the Kronecker symbol that returns 1 if the label has the value $m$ and 0 otherwise, $\pmb{a} = \{a_i\}$, $\pmb{b} = \{b_\mu\}$ and $\pmb{c} = \{c_m\}$ are three sets of local fields acting respectively on the visible and hidden layers and on the label variable. $\pmb{w} = \{w_{i \mu}\}$ is the \textit{weight matrix} that models the interactions between visible and hidden layers, and $\pmb{d} = \{d_{m \mu}\}$ is the \textit{label matrix} that represents the interactions between the label and the hidden layer. The structure of the semi-supervised RBM is sketched in Fig.~\ref{fig:training process}-A. 

\subsection{Out-of-equilibrium training}

\begin{figure*}[!t]
    \centering
    \includegraphics[width=0.9\textwidth]{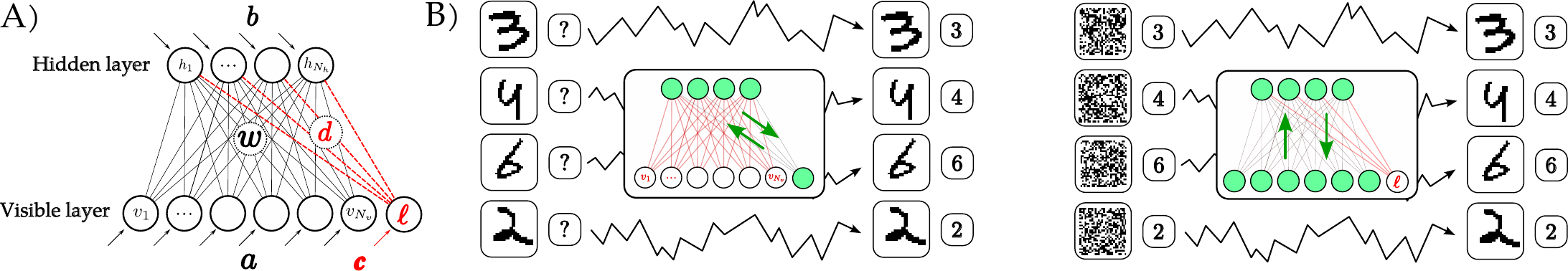}
    \caption{A): Scheme of the semi-supervised RBM. B): Sketch of the sampling procedures used to calculate the two gradients during training. Left): label prediction. The visible layer is clamped to the data, while the labels are initialized randomly. The hidden layer and labels are sampled alternately using block-Gibbs sampling (green) and, after $k$ MCMC steps, the model must provide the correct labels. Right): Conditional Sampling. The labels are fixed and the visible layer is initialized randomly. The model must generate a sample corresponding to the label in $k$ MCMC steps.}
    \label{fig:training process}
\end{figure*}

EBMs are generally trained by maximizing the Log-Likelihood (LL) function of the model computed on the dataset $\mathcal{D} = \{(\pmb{v}^{(1)}, \ell^{(1)}), \dots, (\pmb{v}^{(M)}, \ell^{(M)})\}$
\begin{eqnarray}\label{eq:LL}
    \mathcal{L}(\pmb{\theta} | \mathcal{D}) \!=\! \textstyle \frac{1}{M} \sum_{m=1}^{M} \log p_{\pmb{\theta}}\left(\pmb{v}\! =\! \pmb{v}^{(m)}, \ell = \ell^{(m)}\right)  \nonumber\\ \!=\! \frac{1}{M}\! \sum_{m=1}^{M} \log \sum_{\pmb{h}} e^{- E_{\pmb{\theta}}\left(\pmb{v}^{(m)},\,\pmb{h}, \ell^{(m)}\right)} \!-\! \log Z_{\pmb{\theta}},
\end{eqnarray}
via (stochastic) gradient ascent. As usual, the gradient of $\mathcal{L}$ is obtained by deriving it with respect to all parameters of the model (i.e., $\bm\theta=\{\pmb{a}, \pmb{b},\pmb{c},\pmb{d},\pmb{w}\}$ in our RBMs), which can be written as a subtraction of two terms:
\begin{align}\label{eq:LL gradient}
    \frac{\partial \mathcal{L}}{\partial \theta_i} &= \biggl\langle-\frac{\partial E_{\pmb{\theta}}}{\partial \theta_i}\biggr\rangle_{\mathcal{D}} - \biggl\langle-\frac{\partial E_{\pmb{\theta}}}{\partial \theta_i}\biggr\rangle_{E}.
\end{align}
The symbols $\langle\cdot\rangle_{\mathcal{D}}$, and $\langle\cdot\rangle_{E}$ represent the average over the dataset and the model's Boltzmann measure described in~\eqref{eq:Boltzmann distribution}, respectively. One of the main challenges in training Energy-Based Models (EBMs) is computing a term on the right-hand side of \eqref{eq:LL gradient}, usually estimated via MCMC simulations. This term requires the Markov chains to reach equilibrium—reflecting the Boltzmann measure—before statistical averages can be computed. This process can be very time-consuming, especially with complex datasets. The same issue comes up when generating new data samples according to the Boltzmann distribution. However, as mentioned in the introduction, there is a simple way around this problem~\cite{nijkamp2019learning,decelle2021equilibrium,agoritsas2023explaining}.

The learning dynamics ruled by the gradient in~\eqref{eq:LL gradient} have a fixed point where the (generalized) \textit{moments} of the distribution match those of the dataset, signified by $\langle-\partial E_{\pmb{\theta}}/\partial \theta_i\rangle_{\mathcal{D}} = \langle-\partial E_{\pmb{\theta}}/\partial \theta_i\rangle_{E}$. This indicates that even with accurate gradient computation during training—which is often not achievable—generation with these models is costly. It involves equilibrating the MCMC chains prior to generating good-quality samples. This becomes more challenging as the mixing times increase during training~\cite{decelle2021equilibrium,dabelow2022three}. An alternative approach suggests training the model to replicate the dataset's moments not at equilibrium, but after a few sampling steps, say $k$, from an initial distribution $\bm p_0$. This can be achieved by adjusting the gradient as
\begin{align}\label{eq:OOELLgradient}
    \frac{\partial \mathcal{L}^\mathrm{OOE}}{\partial \theta_i} &= \biggl\langle-\frac{\partial E_{\pmb{\theta}}}{\partial \theta_i}\biggr\rangle_{\mathcal{D}} - \biggl\langle-\frac{\partial E_{\pmb{\theta}}}{\partial \theta_i}\biggr\rangle_{p(k,\bm p_0)}.
\end{align}
Here, $p(k,\bm p_0)$ represents the non-stationary distribution of samples generated through an MCMC process that hasn't reached equilibrium. The model trained this way is optimized to generate quality samples when sampled following the exact same procedure (at the fixed point): same update rules, initialization distribution and number of steps. This possibility has been recently proven rigorously~\cite{agoritsas2023explaining}, and experimentally validated in several studies across different EBMs~\cite{nijkamp2019learning,Nijkamp_Hill_Han_Zhu_Wu_2020,muntoni2021adabmdca}, including RBMs~\cite{decelle2021equilibrium}.

\begin{figure*}[!t]
    \centering
    \includegraphics[width=0.8\textwidth]{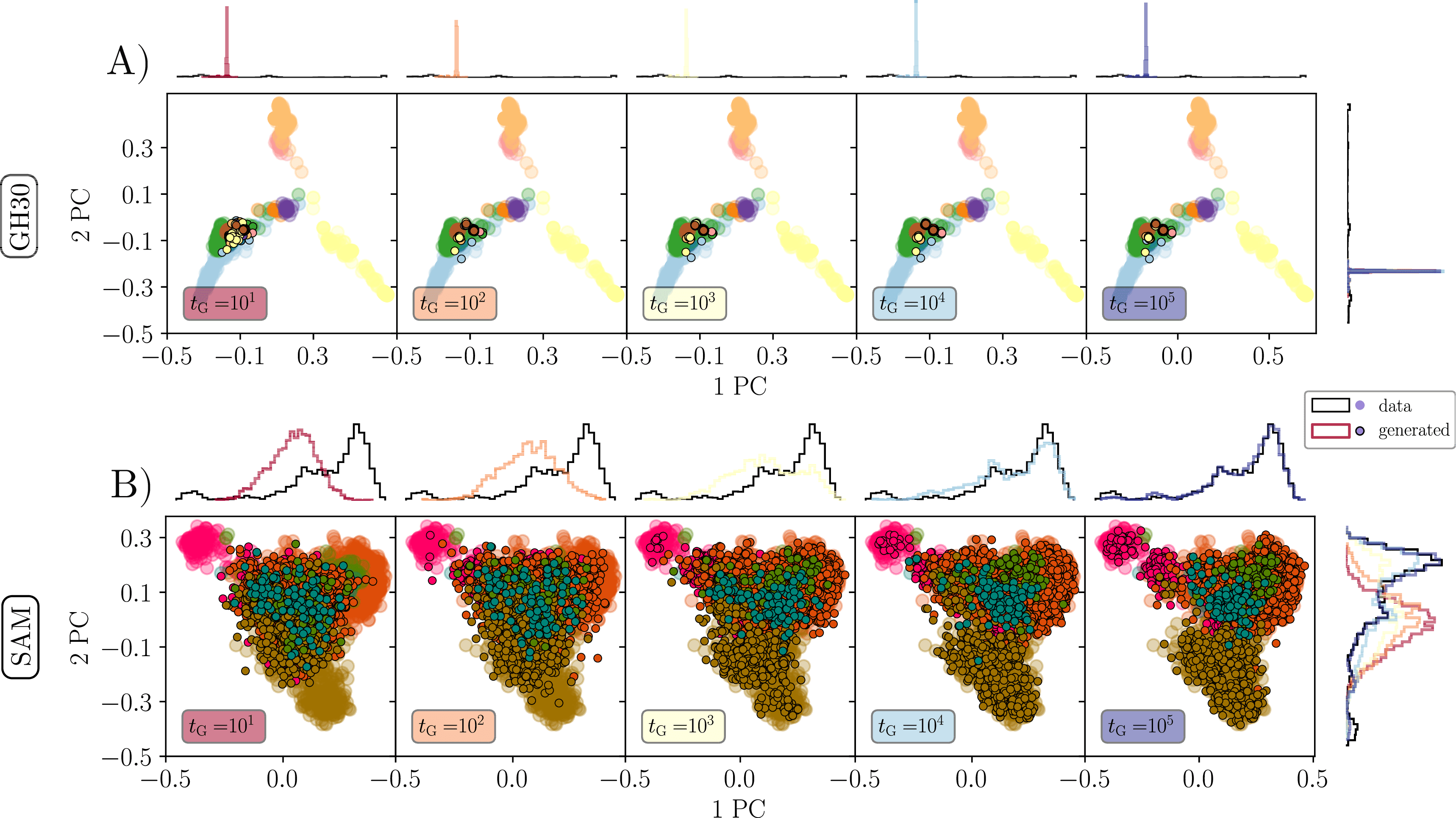}
    \caption{Difficulties in generation with RBMs trained with the PCD-100 protocol. In both panels, we show from left to right the projection on the first two main directions of the dataset of generated samples conditioned on a given label after a different number of sampling steps $t_\mathrm{G}=10,\ 100,\ 1000,\ 10^4$ and $10^5$ respectively. As in Fig.~\ref{fig:PCAs hist Rdm}, each point represents a sample, the labels are shown in different colors, and the synthetic data are highlighted by an outer black ring. In the lateral margins, we show the histograms of the projections along each of the two directions: in black the dataset and the colors refer to the samples generated at different sampling times $t_\mathrm{G}$. In A) we show the results for a dataset where PCD training was unstable, the GH30 dataset. Even up to $t_\mathrm{G}=10^5$, the sampling suffers from strong mode collapse. In B) we show data obtained when training the SAM dataset, where the PCD-100 training leads to a good generative model. We see that in this case, good quality samples that reproduce the statistics of the dataset are generated only after $10^5$ MCMC steps.}
    \label{fig:generation vs sampling time}
\end{figure*}

\begin{figure*}[!t]
    \centering
    \includegraphics[width=0.7\textwidth]{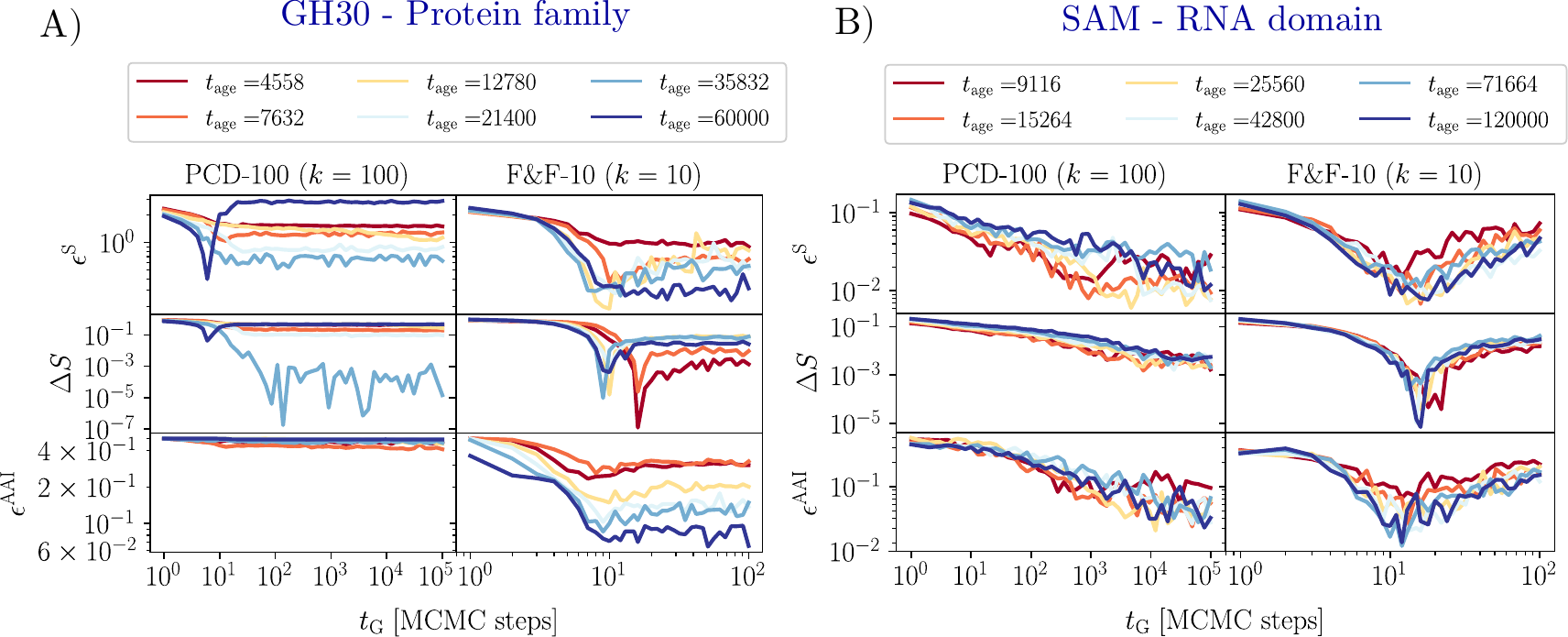}    \caption{Comparison of the scores on the generated data between PCD-100 and F\&F-10 RBMs as a function of the generation time ($t_{\mathrm{G}}$) for A) GH30 and B) SAM. All the scores are computed by comparing the test set with an identical (in terms of samples for each category) generated dataset. The samples of each category of the dataset have been compared with the corresponding samples of the synthetic data, and the curves shown in the figure represent the average scores across the different categories. The different colours of the curves represent different training times ($t_{\mathrm{age}}$), expressed in terms of gradient updates. Notice that for the PCD-RBM the generation time ranges up to $10^5$ MCMC updates, while for the F\&F-RBM it only reaches $10^2$ MCMC updates. The generated samples shown in Figs.~\ref{fig:PCAs hist Rdm} and \ref{fig:generation vs sampling time} correspond to the darkest blue curves in correspondence of the indicated generation time $t_{\mathrm{G}}$.}
    \label{fig:scores comparison}
\end{figure*}

In this paper, we will go one step further. We want to train the RBM to perform not one but two different generative tasks after only $k=10$ MCMC sweeps by manipulating the chain initializations $\bm p_0$. Specifically, we want to train the model to both synthesize (from random) samples of a given label, and to infer the correct label when given a dataset sample as chain initialization. To this end, we use two different out-of-equilibrium gradients in training, each designed for one of these tasks. The difference between the two corresponds to the term $\av{\cdot}_{p(k,\bm p_0)}$ in \eqref{eq:OOELLgradient}. For label prediction, this term is computed by clamping the visible layer onto the images/sequences in the training data and letting evolve the label configurations. For conditional generation, this term is computed using chains where the visible layer is randomly initialized and the labels are kept fixed to the labels in the training data.
A sketch of the sampling procedures used to compute the two gradients can be found in the two panels of Fig.~\ref{fig:training process}-B (label prediction, Fig.~\ref{fig:training process}-B left; data generation, Fig.~\ref{fig:training process}-B right).
We refer to the models trained in this way as F\&F-10 RBMs, where the 10 represents the number of MCMC steps $k$ used to estimate the gradient at each parameter update. The model and hyperparameters used for the training are listed in Tab.~S2 of the SI.

For comparison, we will also train our RBMs using the standard PCD recipe~\cite{tieleman2008training} with a large number $k=100$ of sampling steps by gradient update to try to promote thermalization of the Markov chains as much as possible during training. We will refer to this training procedure as PCD-100. In this training scheme, both the variables and the labels evolve together during sampling, and the last configurations reached after each parameter update are used as initialization for the chains of the subsequent one, which is commonly referred as the permanent or persistent chain.
Furthermore, we train the F\&F-10 and PCD-100 RBMs under the same conditions: same training sets, training epochs, learning rate, and mini-batch size. Since the PCD-100 requires 10 times more sampling steps per update, it is important to emphasize that PCD training takes 10 times longer than the F\&F.

\section{Results} 
We applied the F\&F RBM to five labelled data sets. First, MNIST~\cite{lecun1998gradient}, which contains images of handwritten digits along with their respective number, so that we can easily introduce the method and visually assess the quality of the conditional generation. Second, the Human Genome Dataset (HGD)~\cite{10002015global}, comprising binary vectors representing a human individual with 1s or 0s indicating gene mutation relative to a reference sequence. Labels here signify the individual's continental origin. Third, the GH30 enzyme protein family dataset, as a benchmark for the model's capability to generate artificial protein sequences having a subfamily-specific biological function trained using natural sequences classified in the CAZy database~\cite{lombard2014carbohydrate}. Forth, the S-adenosylmethionine (SAM) dataset, which consists of homologous RNA sequences of the aptamer domain of the SAMI riboswitch, for which taxonomic classification is known. And finally, the Classical Music Piano Composer (CMPC) datasets, which consist of various piano pieces in MIDI format by different classical composers. These datasets allow us to test our method on real biological data of interest with different levels (and types) of characterization, but also on completely different data such as real musical compositions. Detailed explanations of these datasets are available in SI-1.

{\bf Classification task --} Figs.~\ref{fig:PCAs hist Rdm}-B and C illustrate the label prediction accuracy for the test set over training time using both F\&F and PCD training protocols. In our experiments we find that, with both training schemes, the models are able to reach very good accuracies for all the considered datasets, never below 0.89 and compatible between training schemes. Interestingly, we find that the best performance of the F\&F RBM is not achieved at the same number of MCMC steps used for the training, but grows very rapidly and then stabilizes without displaying any out-of-equilibrium effects. We show the prediction accuracy over sampling time $t_\mathrm{G}$ for both training protocols in Fig.~S3 of the SI. We would like to emphasize that our PCD-RBMs perform well for the label prediction task even in cases where they are not able to conditionally generate new samples, as we will show later, consistently with the good classification results obtained in Ref.~\cite{larochelle2012learning}. The confusion matrices from label prediction for all datasets are gathered in the SI, Fig.~S1.

We would also like to emphasize that we observe that one does not need to explicitly train the RBMs to predict the labels in order to obtain a reasonably good prediction, since machines trained with only the conditional generation gradient can also classify the samples. However, we find that the second label gradient significantly increases the accuracy of this classification task, as RBMs trained with two gradients achieve higher label prediction accuracies. The data to support this claim are shown in Supplementary Figures S7 and S8.

{\bf Conditioned Generation task --} We show in Fig.~\ref{fig:PCAs hist Rdm}-A a projection of the samples generated after just 10 MCMC steps with a given label onto the first two principal directions of each dataset using the F\&F. The F\&F model effectively generates data within a few MCMC steps that satisfy the target labels and cover the entire data space following a very similar distribution to the original data, as can be seen from the comparison of the histograms in the figure. 
In the case of the PCD-100 RBM, the situation is somewhat different. First, we find that the trainings are quite unstable depending on the hyperparameters used. For example, by changing the number of hidden nodes we get RBMs that are either good or completely unusable generative models when adopting the same training scheme, as we discuss this in Fig.~S5 of SI for the SAM dataset. Moreover, we were not able at all to properly obtain generative PCD-100 RBMs for the MNIST and GH30 datasets, as we illustrate in Fig.~\ref{fig:generation vs sampling time}--A. The figure shows the projection of the samples generated by the PCD-100 RBM for the protein sequence dataset, for which the Markov chains remain trapped in a very small region of the data space and the machine never generates reliable sequences within $10^5$ MCMC steps. Conversely, the time required to generate sufficiently diverse samples when a working PCD RBM is found, as in the RNA dataset, is extremely long, see Fig.~\ref{fig:generation vs sampling time}--B, where the Acktinomycetota sequences (pink cluster top left) are generated only after $10^5$ MCMC  steps. This is to be compared with the 10 MCMC required with the F\&F-10 RBM shown in Fig.~\ref{fig:PCAs hist Rdm}--A (which also requires 10 times less time to train).

To further assess the generated data's quality, we used several error scores comparing synthetic and real data properties over the sampling time. These scores examine error in the covariance matrix spectrum, $\epsilon^\mathrm{S}$, diversity via an entropy measure, $\Delta S$, and mode collapse and overfitting using the Adversarial Accuracy Indicator~\cite{yale2020generation}, $\epsilon^\mathrm{AAI}$. In all five
cases, the score is always positive and the perfect generation corresponds to an error of zero. Detailed definitions are found in SI~3. In Fig.~\ref{fig:scores comparison}, we show the evolution of these scores (for datasets GH30 and SAM) obtained when sampling configurations starting from random conditions using PCD-100 and F\&F-10 RBMs, plotted as a function of the MCMC generation sampling time ($t_\mathrm{G}$). The results for the MNIST and HGD datasets are shown in Fig.~S4 at the SI. As expected, the F\&F-10 RBMs develop a best-quality peak around $t_\mathrm{G}\sim 10$, the $k$ used for training, while the PCD-100 runs show a very slow relaxation behaviour eventually reaching good quality values at long sampling times in the case of the HGD and SAM datasets, as discussed in Fig.~\ref{fig:generation vs sampling time}--B, or show a clear problem in generative performance with very poor scores (MNIST and GH30 datasets) and bad generated samples, as shown in the projections of Fig.~\ref{fig:generation vs sampling time}--A and visually in Fig. S2 in the SI for MNIST.

Interestingly, based on previous experience with these datasets without label monitoring, we found the poor performance of the semi-supervised PCD-100 on MNIST training very surprising, since PCD usually performs quite well on this dataset (see, for example, \cite{decelle2021equilibrium}). The only explanation we have is that the addition of labels seems to greatly increase the mixing times, as can be seen in Fig. S4--A in the SI. A similar situation is found with the SAM dataset when 1000 hidden nodes are used, see Fig. S5--B in the SI, but not with 100 hidden nodes. It is possible that a working setting can also be found for the MNIST dataset, but we did not succeed in our trials. In any case, this difficulty may explain why previous similar approaches that addressed classification using RBMs~\cite{larochelle2012learning} focused only on the classification task and did not consider the conditional generation performance. Conversely, even though the HGD is typically a challenging benchmark dataset for classical equilibrium RBM models \cite{10.21468/SciPostPhys.14.3.032}, our semi-supervised training yielded very high-quality models with thermalization times of only a few hundred MCMC steps.  
Altogether, these results show that classical training of RBMs with PCD for conditional generation is unreliable because it is difficult to control the risks of very poor generation performance. In contrast, the F\&F model proved to be robust and reliable for all tested highly structured datasets. It produced high-quality artificial data after only a few MCMC steps, while requiring a significantly shorter training time.
\begin{figure*}[!t]
    \centering
    \includegraphics[width=0.8\textwidth]{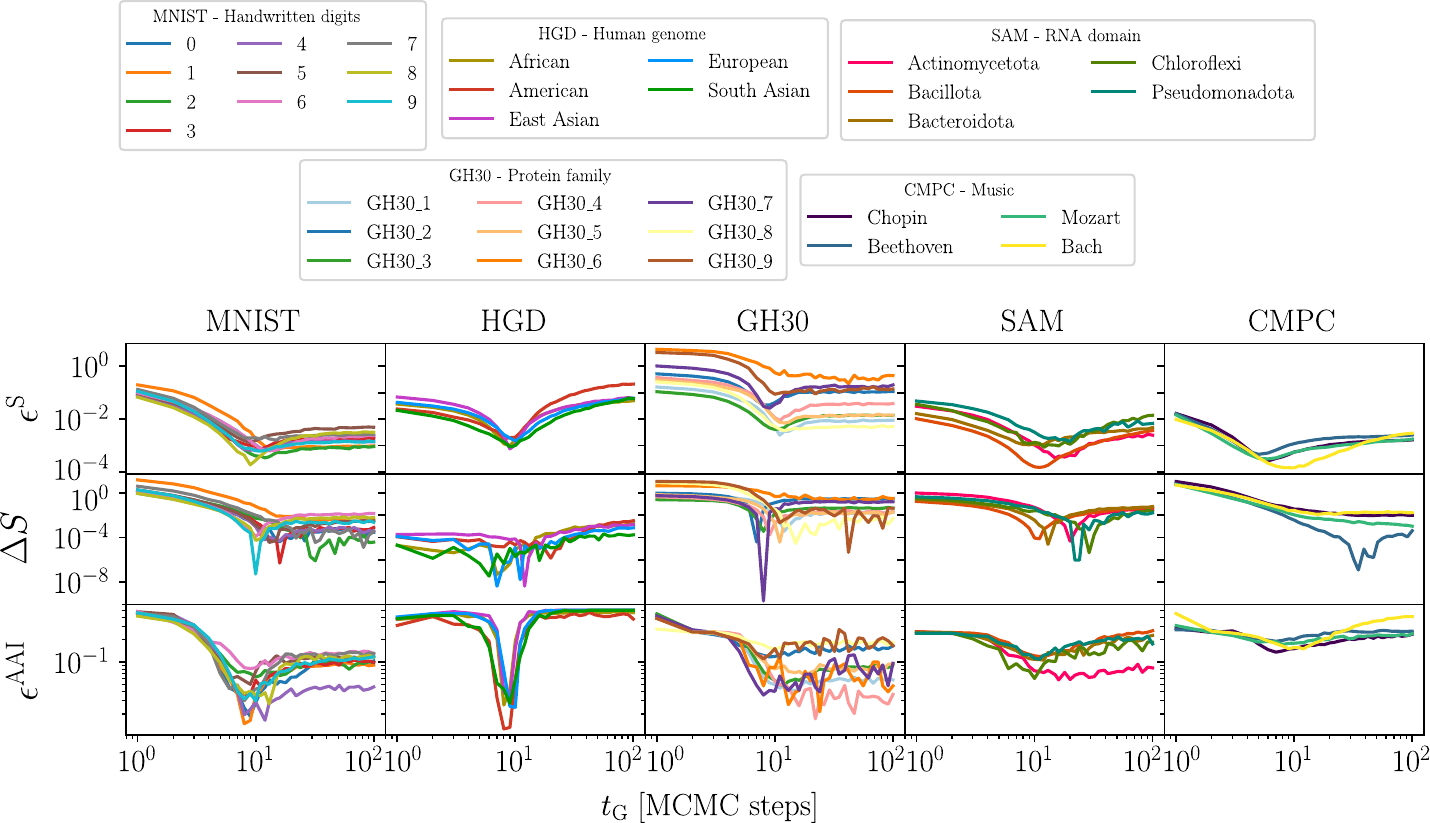}
    \caption{For each of the datasets considered, we show the evolution of three different quality scores as a function of sampling generation time, $t_{\mathrm{G}}$, for each label separately. The first row shows the error on the eigenvalue spectra, the second row shows the error on the entropy, and the third row shows the Adversarial Accuracy Indicator. For the GH30 and the SAM datasets, we used the training set to generate the error curves because there was too limited data in the test set to compare certain categories. The definition of the scores can be found in the SI~3.}
    \label{fig:scores vs time}
\end{figure*}

We can also examine the quality of the samples generated by the F\&F-10 RBM, but this time for each label, see Fig.~\ref{fig:scores vs time}. Again, the highest quality samples of each category are generated in about 10 steps, the same number of steps used for gradient estimation during training. It is interesting to point out that the F\&F-10 RBM generates good quality samples even for categories in which it has seen only a few examples ($\lesssim 100$) in the training set. The number of samples in the training/test sets for each dataset and category can be found in Table S1.

\begin{figure}[!t]
    \centering
    \includegraphics[width=0.35\textwidth]{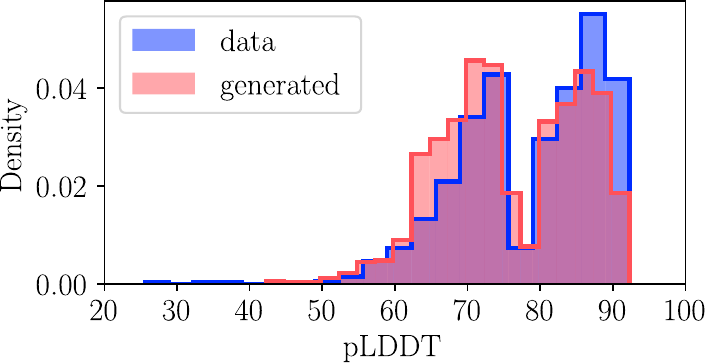}    \caption{Histograms of
    the confidence on the predicted folding structure of GH30 (based on the pLDDT score) for generated data (red) and real data (blue). The generated set consists of 150 samples per each of the 9 categories.}
    \label{fig:GH30 pLDDT}
\end{figure}

For an indirect and more biologically relevant measure of generated protein sequences' quality, we extensively assessed their predicted three-dimensional structure using the \texttt{ESMFold}~\cite{lin2023evolutionary}, comparing these predictions with those obtained with the test set. Specifically, we created histograms for both the generated sequences and the test set based on the frequency of predicted pLDDT scores from \texttt{ESMFold}, indicating the average confidence in the folding. Given a reference protein structure and a structure predicted by a model, the LDDT (Local Distance Difference Test) score assesses how well local atomic interactions in the reference protein structure are reproduced in the prediction. The pLDDT (predicted LDDT) score is returned by the \texttt{ESMFold} model, and it allows us to evaluate the degree of confidence of a folding even without having the reference structure.
These distributions are displayed in Fig.~\ref{fig:GH30 pLDDT} showing a remarkable agreement supporting the idea that we are generating structurally reliable and biologically relevant protein sequences but experimental tests are needed to confirm the biological classification.

As in the classification task, in the conditional generation task we can investigate whether label prediction is beneficial by considering a single gradient (only for conditional generation) or two gradients (both label and sample generation). As can be seen in Fig. S6, we cannot see any particular improvement (nor deterioration) in the quality of the generated samples when we consider two gradients compared to just one. This makes sense, because the information of the label is already contained in the gradient for the conditional generation.

\section{Conclusions}
In this study, we used a unique method for training RBMs to embed the statistics of the datasets into the nonstationary distributions of a Markov chain process~\cite{nijkamp2019learning,decelle2021equilibrium,agoritsas2023explaining}, in contrast to conventional methods that encode information only at the equilibrium measure level. This strategy allows us to use RBMs as efficient generators, similar to diffusion models, with the added benefit that various generative tasks can be easily encoded into the model. In particular, we trained RBMs to generate label-conditioned samples in a minimal number of sampling steps-- a process that is typically tedious and slow in conventional methods-- and derive the good label when Markov chains are randomly initialized. We have shown that our approach successfully generates high-quality synthetic samples that accurately reflect the full diversity of the dataset even for highly structured data, overcoming the limitations of standard (equilibrium) training methods while requiring shorter training times and displaying better stability. Noticeably, our method proved to be able to generate artificial protein sequences covering the full data space across subfamilies with different functionalities, a particularly challenging task. Last but not least, the two-gradient method presented here can be easily implemented in EBMs having a more elaborated energy function (convolutional layers, ...) to model other complex datasets.

\section{Acknowledgements}
We acknowledge financial support by the Comunidad de Madrid and the Complutense University of Madrid (UCM) through the Atracción de Talento programs (Refs. 2019-T1/TIC-13298 and 2019-T1/TIC-12776), the Banco Santander and the UCM (grant PR44/21-29937), and Ministerio de Econom\'{\i}a y Competitividad, Agencia Estatal de Investigaci\'on  and Fondo Europeo de Desarrollo Regional (Ref. PID2021-125506NA-I00).

\section*{Code availability}
The python code for the present work can be found at the link \href{https://github.com/DsysDML/FeF.git}{https://github.com/DsysDML/FeF.git}.

\section*{Biography}

\begin{IEEEbiography}[{\includegraphics[width=1in,height=1.25in,clip,keepaspectratio]{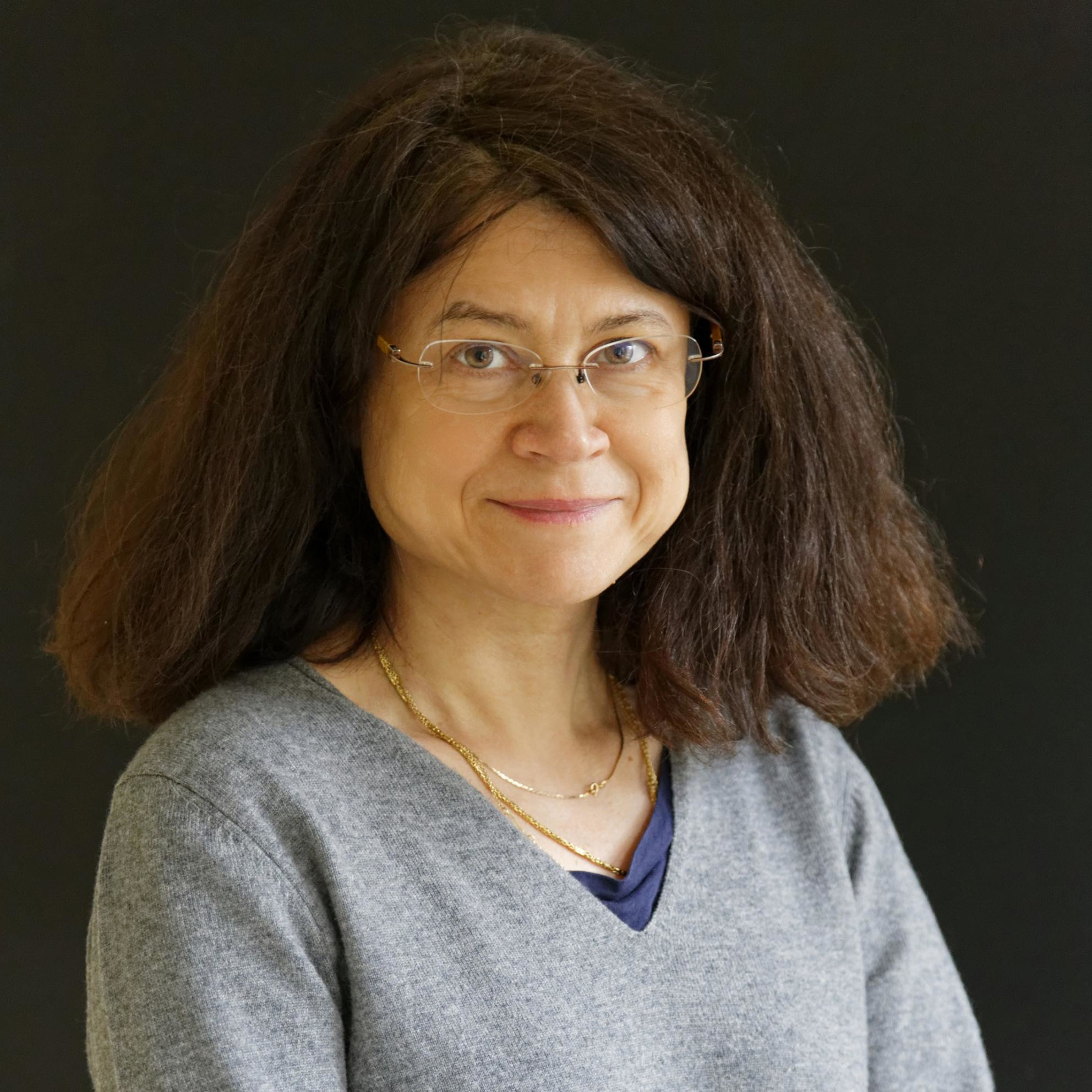}}]{Alessandra Carbone}
is Distinguished Professor of Computer Science at
Sorbonne Université. She has led the Analytical Genomics team since 2003 and has been the director of the Department of Computational and Quantitative Biology since 2009. Her research group addresses computational challenges in the functioning and evolution of biological systems, developing and applying mathematical methods in statistics, deep learning and combinatorial optimization to investigate fundamental principles of cellular functioning, starting from genomic and structural data. Her projects focus on understanding evolution and co-evolution of molecular structures within the cell, encompassing genome-wide  sequence and protein evolution. AC is a senior member of the Institut Universitaire de France (2013-18, 2023-28), and has been recognized with the Prix Joliot-Curie in 2010 and the Grammaticakis-Neuman Prize from the French Academy of Sciences in 2012.
\end{IEEEbiography}

\begin{IEEEbiography}[{\includegraphics[width=1in,height=1.25in,clip,keepaspectratio]{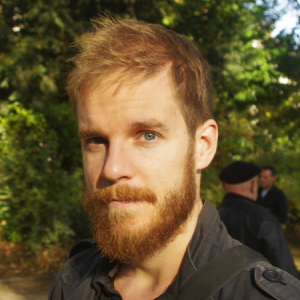}}]{Aurélien Decelle}
is a researcher specialized in Statistical Physics, at the interface between disordered systems and Machine Learning. His works have been covering Bayesian Inference problems such as community detection or Gaussian mixtures as well as Machine Learning problems studying on one hand the phase diagram and learning aspects of Boltzmann and Restricted Boltzmann Machine while also applying generative modelling to Cosmology. He has a Junior PI position at Universidad Complutense de Madrid and is co-head of the group of disorder systems of the theoretical physics department.
\end{IEEEbiography}

\begin{IEEEbiography}[{\includegraphics[width=1in,height=1.25in,clip,keepaspectratio]{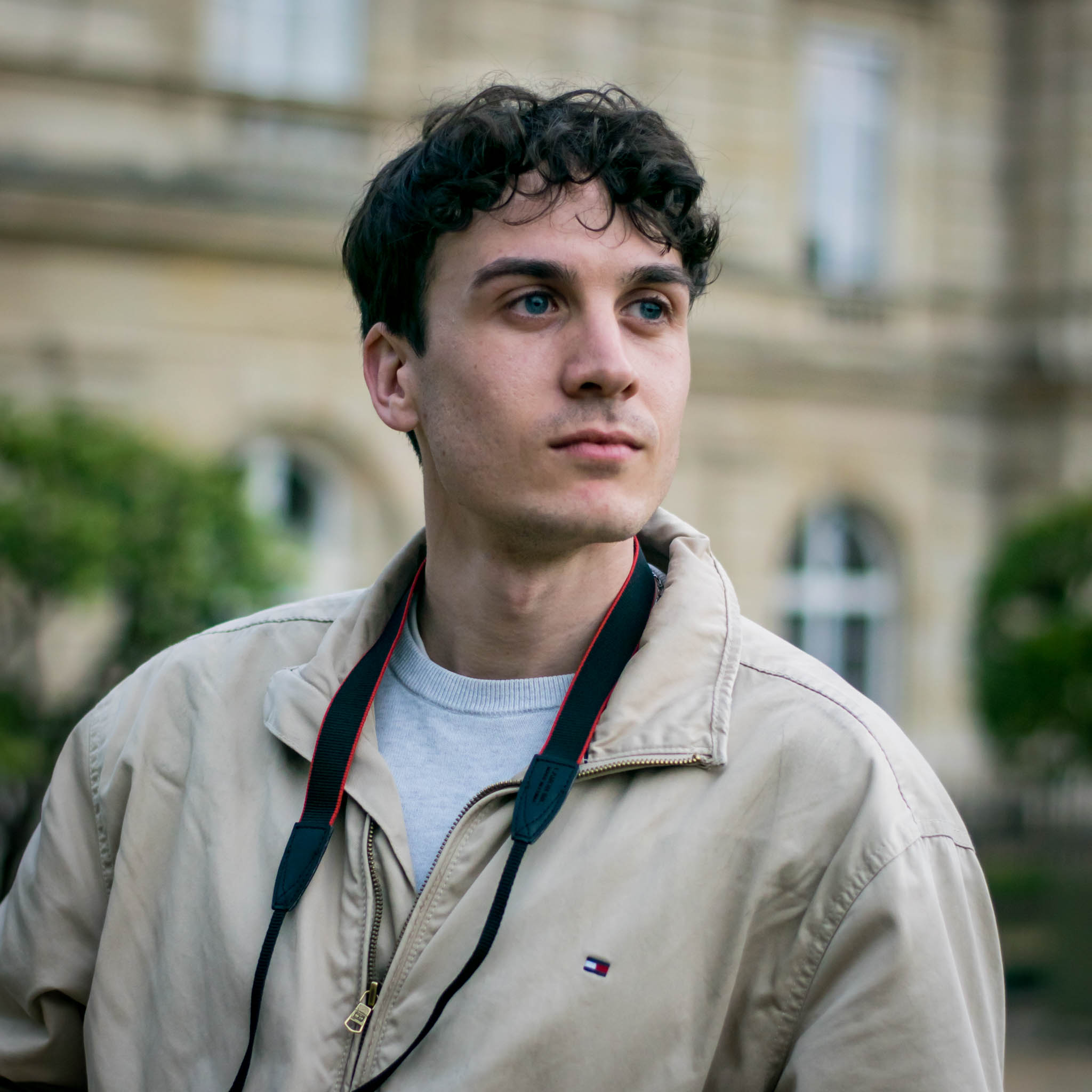}}]{Lorenzo Rosset}
is a PhD student at the École Normale Supérieure in Paris, collaborating with Sorbonne University in Paris and "La Sapienza" University of Rome. Before his PhD, he worked as a researcher at the Complutense University of Madrid. He graduated in “Physics of Data” at Padova University (Italy), where he also obtained his bachelor’s degree in Physics.
His research focuses on generative machine learning models and their application to biological data, specifically the Direct Coupling Analysis (DCA) methods and the Restricted Boltzmann Machine.
\end{IEEEbiography}

\begin{IEEEbiography}[{\includegraphics[width=1in,height=1.25in,clip,keepaspectratio]{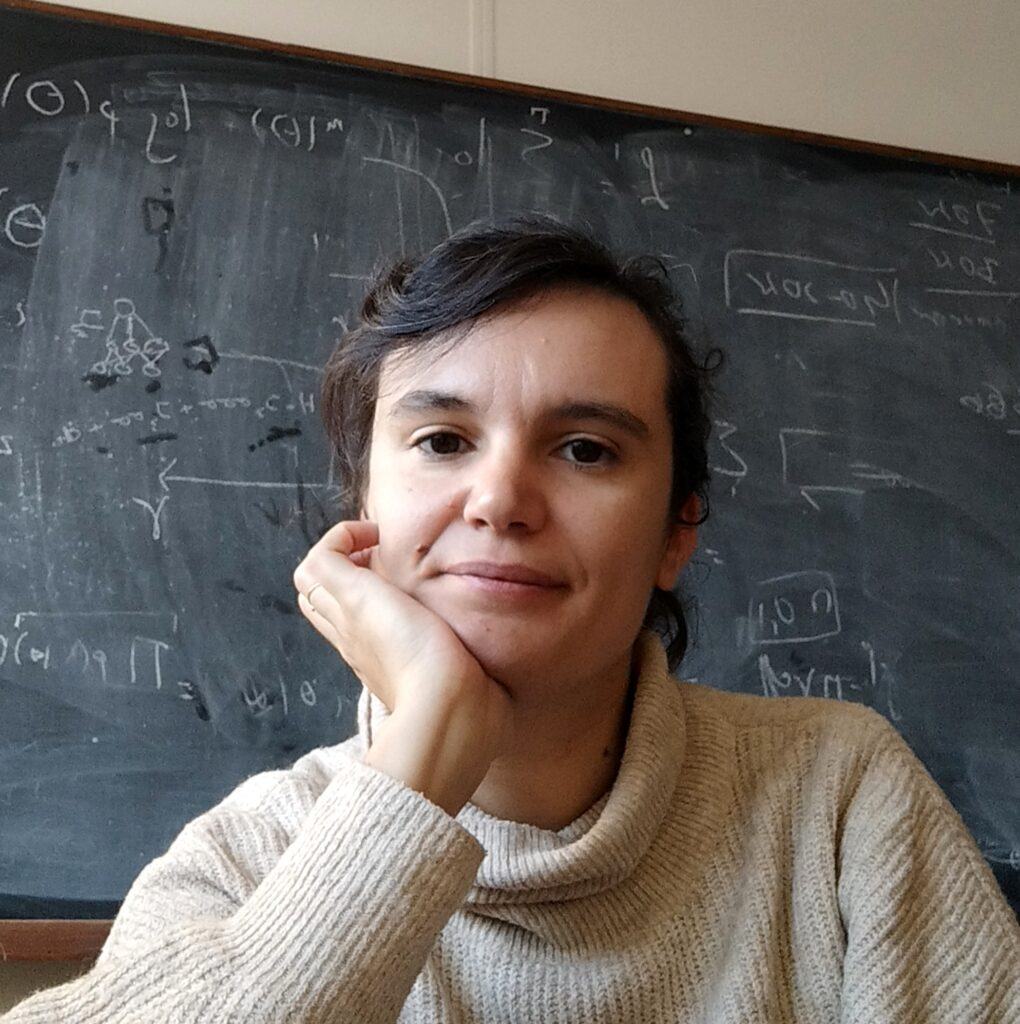}}]{Beatriz Seoane}
is an interdisciplinary researcher specializing in statistical and computational physics. Her work mainly involves numerical studies of disordered systems such as glasses, proteins or neural networks, as well as the development of computer tools to simulate them. She currently holds a permanent position in the Department of Theoretical Physics department at the Complutense University of Madrid. In her research, she applies principles of statistical mechanics to improve the training and interpretability of generative machine learning models and she has specialized in Restricted Bolzmann Machines. In recent years, she has also led several research projects combining physics and artificial intelligence. Previously, she held a junior professorship in “Physics for Machine Learning” at the Faculty of Computer Science of the Université Paris Saclay and completed research stays at several universities across Europe, including the Sorbonne University in Paris, the École Normale Supérieure in Paris and "La Sapienza" University of Rome.
\end{IEEEbiography}

\bibliographystyle{plain}
\bibliography{bibliography}

\newpage
\onecolumn
\section*{Supplementary Information}

\section{Dataset description}
\label{Dataset description}

\subsection{MNIST dataset}
The MNIST dataset~\cite{lecun1998gradient} consists of $28\times 28$ grayscale images of handwritten digits tagged with a label indicating the digit represented, from 0 to 9. We first extracted a training set and a test set of respectively 10000 and 2000 images, and we then binarized the data by setting each pixel to 1 if the normalized value was above 0.3, and to 0 otherwise. To be fed to the RBM, the images have to be flattened into 784-dimensional binary vectors.

\subsection{Human Genome Dataset (HGD)}
The Human Genome dataset (HGD) ~\cite{10002015global} represents the human genetic variations of a population of 5008 individuals sampled from 26 populations in Africa, East Asia, South Asia, Europe, and the Americas. Each sample is a sequence of 805 binary variables, $v_i \in \{0, 1\}$, representing the change alteration or not of a gene relative to a reference genetic sequence. Sequences are classified based on the continental origin of individuals. We trained the RBM on 4507 samples and retained 501 samples for the test set.

\subsection{GH30 family}
The glycoside hydrolases (EC 3.2.1.-), GH for short, are a family of enzymes that hydrolyze the glycosidic bond between two or more carbohydrates or between a carbohydrate and a non-carbohydrate moiety. GH30 is one of the GH families that has been divided into subfamilies in \href{http://www.cazy.org/GH30.html}{CAZy}. It includes nine different subfamilies (GH30-1,..., GH30-9) corresponding to 11 different enzymatic chemical reactions. We created a training and test set of respectively 3922 and 975 annotated sequences from CAZy~\cite{lombard2014carbohydrate,cantarel2009carbohydrate}, having care of reproducing the same samples-per-label proportion between training and test sets. The sequences were previously aligned in an MSA matrix using the MUSCLE algorithm \cite{edgar2004muscle} with default parameters. We then cleaned all MSA columns in which the proportion of gaps was above 70\% of the entries. The resulting sequences have a length of $N_v=430$, where each residue can take one over 21 possible values (20 amino acids + the alignment gap).

\subsection{SAM domain}
The SAM (S-Adenosyl methionine) riboswitch (RF00162) is found upstream of a number of genes which code for proteins involved in methionine or cysteine biosynthesis in Gram-positive bacteria. We downloaded the dataset from \hyperlink{https://rfam.org/}{Rfam} alongside the taxonomic classification of each sequence. Among all the sequences, we retained only those belonging to the five largest groups containing more than 100 sequences: Actinomycetota, Bacillota, Bacteroidota, Chloroflexi and Pseudomonadota. We then split the dataset into training and test set of respectively 4733 and 524 samples having the same proportion of data points across the categories. The aligned sequences have a length of $N_v=108$, and each site can assume one of 5 possible states (the 4 nucleobases + the alignment gap).

\subsection{Classical Music Piano Composer}
This dataset is made of four classical piano pieces in the MIDI format, the Nocturne Op. 27 No. 2 by Chopin, the Piano Sonata ``Pathetique'' Op. 13 - II by Beethoven, the Piano Sonata nº12, K. 332 second movement by Mozart and the Toccata BMV 913 by Bach. For each piece, the MIDI score was turn into an array where each sample corresponds to a given time window, indicating which notes of the piano have been played (by using a one) or not (using a zero). Since the timeframe is quite short and hence the same piano's keys are repeated over many timeframes. The datasets have therefore binned convoluted with a Heaviside function such that two consecutive timeframes could not be identical, for each timeframe a piano's key can either be zero or one, and keeping the number of piano's key of order $\sim 4,5$. Finally, in order to retain the temporal order of the piece, the dataset was made of various consecutive timeframes $T$ put together. Hence, the dimension of input is $N_v = 88 T$, where $88$ is the number of keys on a piano.

The details about the composition of the training/testing sets used for each dataset can be found in Table~\ref{tab:categories count}.
\label{Tables}
\begin{table*}[!t]
    \centering
    \caption{Number of data samples for each category in the train and test sets for the used datasets.}
    \begin{tabular}{ccccccccccc}
        &&&&&MNIST&&&&& \\
        \textbf{Label}&0&1&2&3&4&5&6&7&8&9 \\
        \midrule
        \textbf{Train set count}&1022&1078&1046&1031&965&916&972&1042&977&951
        \\
        \midrule
        \textbf{Test set count}&188&224&218&191&220&174&208&178&197&202 \\
        \bottomrule
    \end{tabular}
    
\vspace{0.5cm}

\begin{tabular}{cccccc}
    &&&HGD&& \\
    \textbf{Label} & African & American & East Asian & European & South Asian \\
    \midrule
    \textbf{Train set count} & 1184 & 622 & 912 & 910 & 879 \\
    \midrule
    \textbf{Test set count} & 138 & 72 & 96 & 96 & 99 \\
    \bottomrule
\end{tabular}

\vspace{0.5cm}

    \begin{tabular}{cccccccccc}
        &&&&&GH30&&&& \\        \textbf{Label}&GH30\_1&GH30\_2&GH30\_3&GH30\_4&GH30\_5&GH30\_6&GH30\_7&GH30\_8&GH30\_9 \\
        \midrule
        \textbf{Train set count}& 886 & 287 & 1044 & 270 & 435 & 39 & 89 & 810 & 62 \\
        \midrule
        \textbf{Test set count}& 221 & 71 & 260 & 67 & 108 & 9 & 22 & 202 & 15 \\
        \bottomrule
    \end{tabular}

    \vspace{0.5cm}

\begin{tabular}{cccccc}
    &&&SAM&& \\
    \textbf{Label} & Actinomycetota & Bacillota & Bacteroidota & Chloroflexi & Pseudomonadota \\
    \midrule
    \textbf{Train set count} & 549 & 3159 & 747 & 120 & 158 \\
    \midrule
    \textbf{Test set count} & 61 & 351 & 82 & 13 & 17 \\
    \bottomrule
\end{tabular}

\vspace{0.5cm}
\begin{tabular}{cccccc}
    &&CMPC&& \\
    \textbf{Label} & Chopin & Beethoven  & Mozart & Bach \\
    \midrule
    \textbf{Train set count} & 4427 & 3716 & 3716 & 5973 \\
    \midrule
    \textbf{Test set count} & 1106 & 929 & 929 & 1493 \\
    \bottomrule
\end{tabular}
\label{tab:categories count}

\end{table*}

\section{RBM training details}\label{RBM details}
The hyperparameters used for the training processes discussed in this paper are given in Table~\ref{tab:RBMs specifics}.

\begin{table*}[!t]
    \centering
    \caption{Hyper-parameters of the RBMs used in this work.}
    \begin{tabular}{ccccccc}
    \textbf{dataset} & \textbf{epochs} & \thead{\textbf{minibatch} \\ \textbf{size}} & \thead{\textbf{total gradient} \\ \textbf{updates}} & $\boldsymbol{k}$ & \thead{\textbf{learning} \\ \textbf{rate}} & $\boldsymbol{N_\mathrm{h}}$  \\
    \midrule
    MNIST (PCD) & 30000 & 500 & $6 \cdot 10^5$ & 100 & $10^{-2}$ & 1024\\
    \midrule
    HGD (PCD) & 30000 & 4507 & $3 \cdot 10^4$ & 100 & $10^{-2}$ & 1024 \\
    \midrule
    GH30 (PCD) & 30000 & 1961 & $6 \cdot 10^4$ & 100 & $10^{-2}$ & 1024 \\
    \midrule
    SAM (PCD) & 30000 & 1000 & $1.2 \cdot 10^5$ & 100 & $10^{-2}$ & 100 / 1000\\
    \midrule
    CMPC (PCD) & 30000 & 2000 & $2.7 \cdot 10^5$ & 100 & $10^{-2}$ & 500
    \\
    \midrule
    MNIST (F\&F) & 30000 & 500 & $6 \cdot 10^5$ & 10 & $10^{-2}$ & 1024\\
    \midrule
    HGD (F\&F) & 30000 & 4507 & $3 \cdot 10^4$ & 10 & $10^{-2}$ & 1024 \\
    \midrule
    GH30 (F\&F) & 30000 & 1961 & $6 \cdot 10^4$ & 10 & $10^{-2}$ & 1024 \\
    \midrule
    SAM (F\&F) & 30000 & 1000 & $1.2 \cdot 10^5$ & 10 & $10^{-2}$ & 100 / 1000\\
    \midrule
    CMPC (F\&F) & 30000 & 2000 & $2.7 \cdot 10^5$ & 10 & $10^{-2}$ & 1024
    \\
    \bottomrule
    \end{tabular}
    
    \label{tab:RBMs specifics}
\end{table*}

\section{Quality scores}
\label{Quality scores}
To assess the generation capabilities of the RBM, one can compute a set of observables on the generated dataset and the actual data and compare them \cite{decelle2021equilibrium}. In the plots of Figs.~3 and~4 we have considered the following scores:
\begin{itemize}
    \item \textbf{Error on the spectrum ($\mathbf{\epsilon^{S}}$)}: Given a data matrix $X \in \mathbb{R}^{M \times N_v}$, its singular value decomposition (SVD) consists in writing $X$ as the matrix product $$X = U S V^T,$$ where $U \in \mathbb{R}^{M \times M}$, $S$ is an $M \times N_v$ matrix with the singular values of $X$ in the diagonal, and $V \in \mathbb{R}^{N_v \times N_v}$. Let us call $N_s = \min(M, N_v)$. Once we sort the singular values $\{s_i\}$ such that $s_1 > s_2 > \cdots > s_{N_s}$, we can define the error of the spectrum as
    \begin{equation}
        \epsilon^{S} = \frac{1}{N_s} \sum_{i=1}^{N_s} \left(s_i^{\mathrm{data}} - s_i^{\mathrm{gen}}\right)^2,
    \end{equation}
    where $\{s_i^{\mathrm{data}}\}$ are the singular values of the true data and $\{s_i^{\mathrm{gen}}\}$ are the singular values of the generated dataset.
    \item \textbf{Error on the entropy ($\mathbf{\Delta S}$)}: We approximate the entropy of a given dataset by its byte size when compressed with gzip~\cite{baronchelli2005measuring}. In particular, if $S^{\mathrm{data}}$ is the estimated entropy of the true data and $S^{\mathrm{gen}}$ is the estimated entropy of the generated data, we define the error of entropy as \begin{equation} \Delta S = \left( \frac{S^{\mathrm{gen}}}{S^{\mathrm{data}}} - 1\right)^2. \end{equation} 
    A large $\Delta S$ indicates that the generated set lacks diversity or that the generated samples are less ``ordered'' than the dataset.
    \item \textbf{Error on the Adversarial Accuracy Indicator ($\mathbf{\epsilon^{\mathrm{AAI}}}$)}: This score was introduced in Ref.~\cite{yale2020generation} to quantify the similarity and ``privacy'' of data drawn from a generative model with respect to the training set. We first construct a dataset obtained by joining the real dataset with the generated dataset, and then compute the matrix of distances between each pair of data points. We denote by $P_{\mathrm{GG}}$ the probability that a generated datapoint has as the nearest neighbour a generated data and by $P_{\mathrm{ DD }}$ the probability that a true datapoint has as the nearest neighbour a true data. In the best case, when the generated data are statistically indistinguishable from the true ones, we have $P_{\mathrm{GG}} = P_{\mathrm{ DD }} = 0.5$. Therefore, we can define the error of the Adversarial Accuracy Indicator as follows:
    \begin{equation}
        \epsilon^{\mathrm{AAI}} = \frac{1}{2} \left[(P_{\mathrm{GG}} - 0.5)^2 + (P_{\mathrm{DD}} - 0.5)^2 \right].
    \end{equation}
    
\end{itemize}

\section{Supplementary figures}
\label{Supplementary figures}
In this section you will find some additional illustrations referred to in the main text.
\begin{figure*}[!t]
    \centering
    \includegraphics[width=\textwidth]{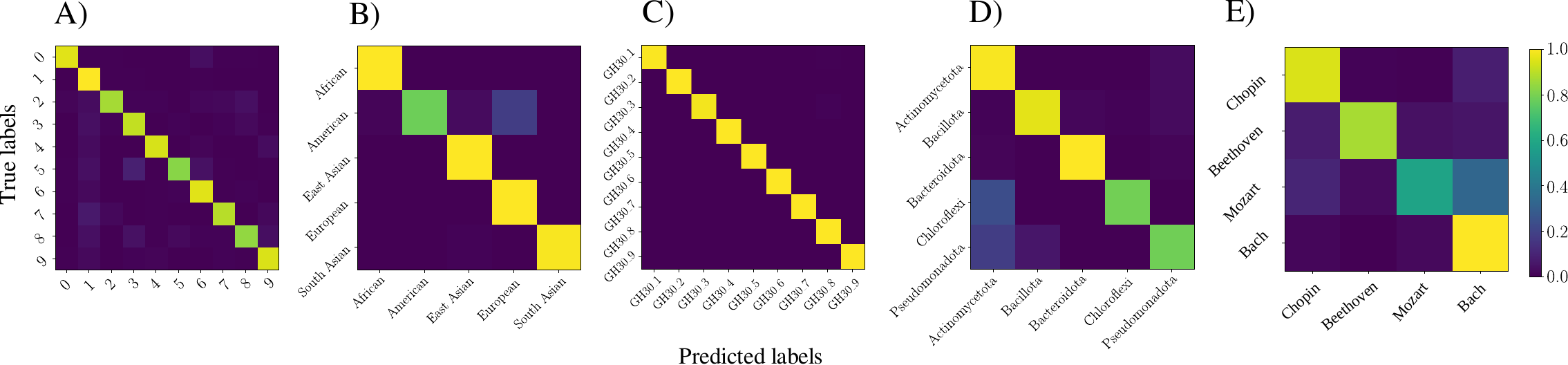}
    \caption{Confusion matrices for the label classification using F\&F on the test sets of A) MNIST, B) HGD, C) GH30, D) SAM and E) CMPC.}
    \label{fig:confusion matrices}
\end{figure*}

\begin{figure*}[!t]
    \centering
    \includegraphics[width=0.8\textwidth]{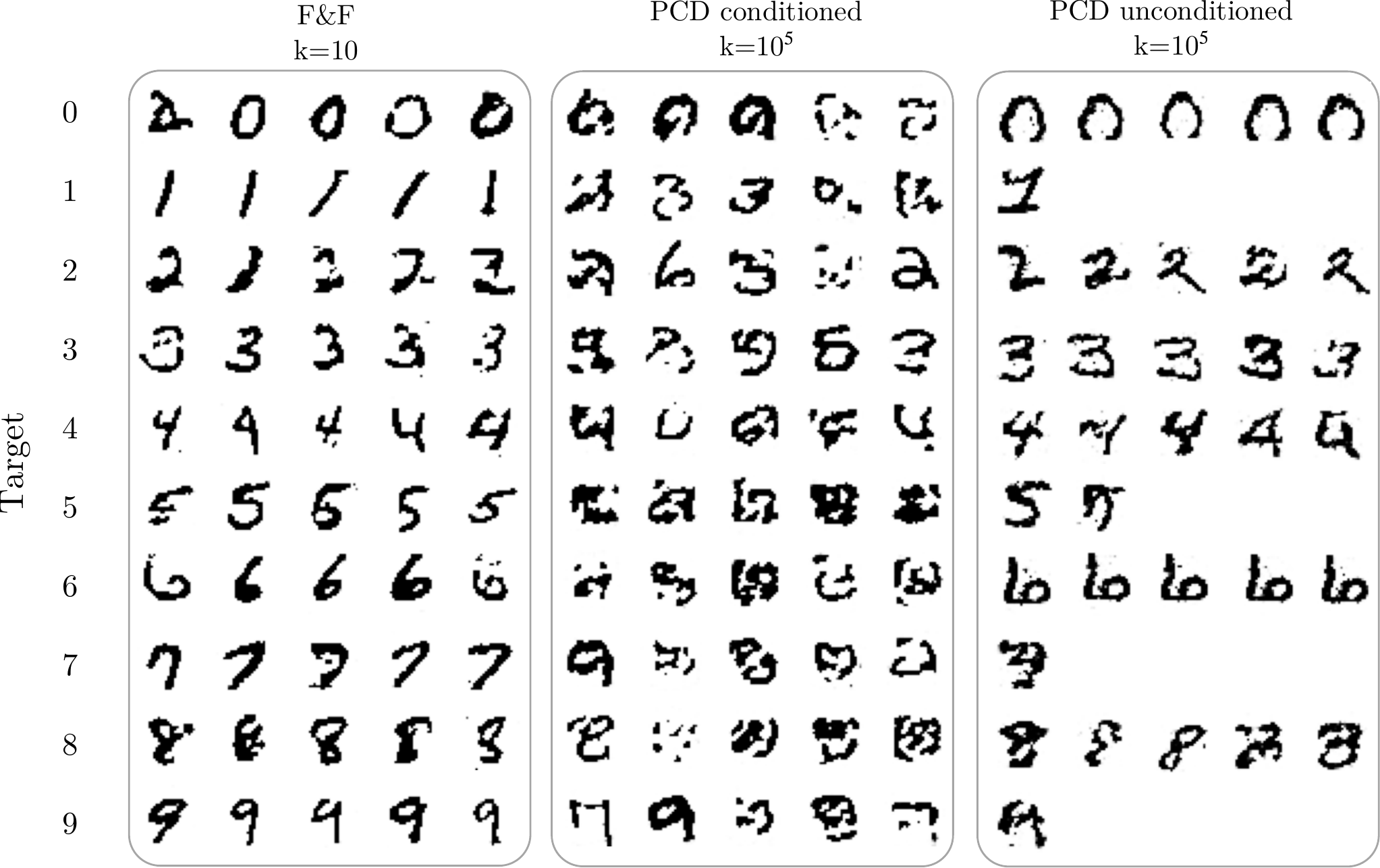}
    \caption{MNIST images created using different methods for specific labels. From left to right, the first box shows the output of F\&F for $k=10$ MCMC steps. The images in the second box are generated using a PCD-RBM after $10^5$ MCMC steps when the Markov chains are clamped to a specific label value. The third box shows the result of sampling with a PCD-RBM, where we also sample the labels when running the Markov chains. An empty slot means that the RBM never provided the appropriate sample in our tests after $10^5$ MCMC steps.}
    \label{fig:MNIST samples}
\end{figure*}

\begin{figure*}[!t]
    \centering
    \includegraphics[width=\textwidth]{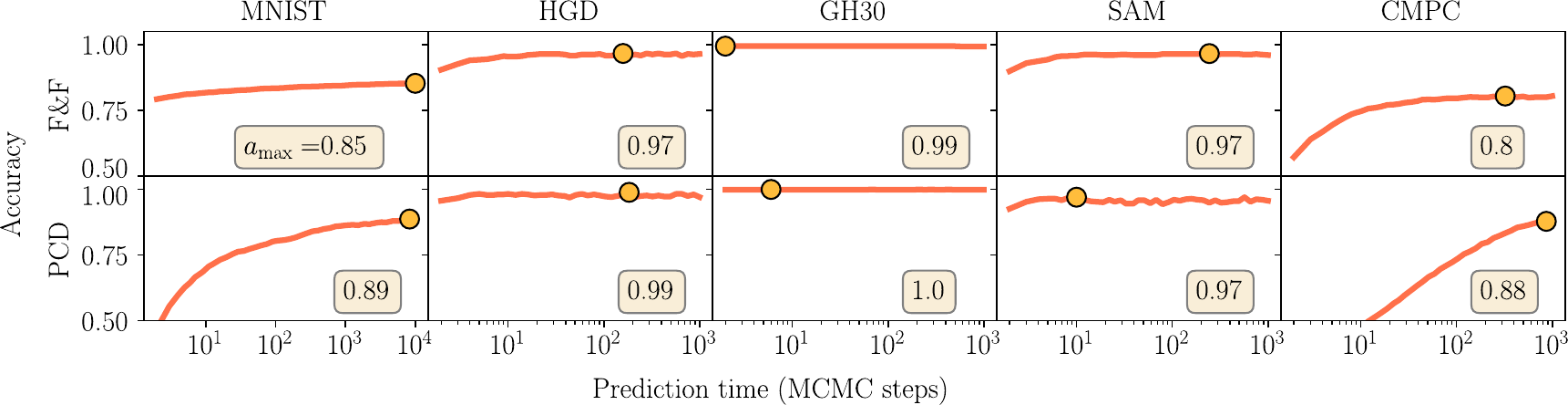}
    \caption{Test accuracies obtained by the best models in the task of label prediction as a function of the prediction time for all the considered datasets. The yellow boxes in the bottom-right corner of the insets indicate the maximum accuracy achieved ($a_{\mathrm{max}}$), corresponding to the big yellow dots.}
    \label{fig:acc vs sampling time}
\end{figure*}

\begin{figure*}[!t]
    \centering
    \includegraphics[width=0.8\textwidth]{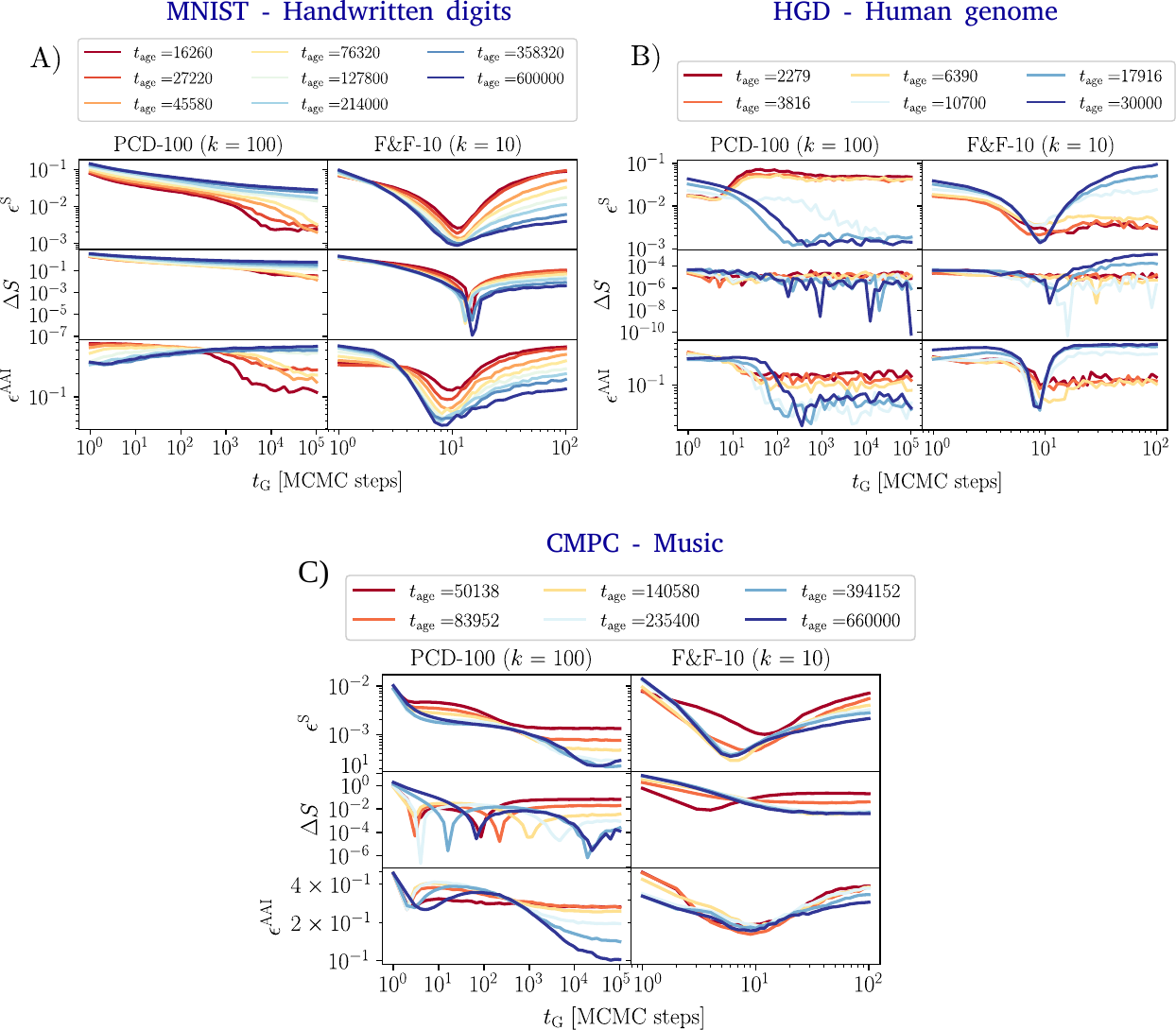}
    \caption{Comparison of the scores on the generated data between PCD-100 and F\&F-10 RBMs as a function of the generation time ($t_{\mathrm{G}}$) for A) MNIST, B) HGD and C) CMPC. All the scores are computed by comparing the test set with an identical (in terms of samples for each category) generated dataset. The samples of each category of the dataset have been compared with the corresponding samples of the synthetic data, and the curves shown in the figure represent the average scores across the different categories. The different colours of the curves represent different training times ($t_{\mathrm{age}}$), expressed in terms of gradient updates. Notice that for the PCD-RBM the generation time ranges up to $10^5$ MCMC updates, while for the F\&F-RBM it only reaches $10^2$ MCMC updates. The generated samples shown in Figs.~1-A and 3 correspond to the darkest blue curves in correspondence of the indicated generation time $t_{\mathrm{G}}$.}
    \label{fig:scores comparison MNIST and HGD}
\end{figure*}

\begin{figure*}[!t]
    \centering
    \includegraphics[width=\textwidth]{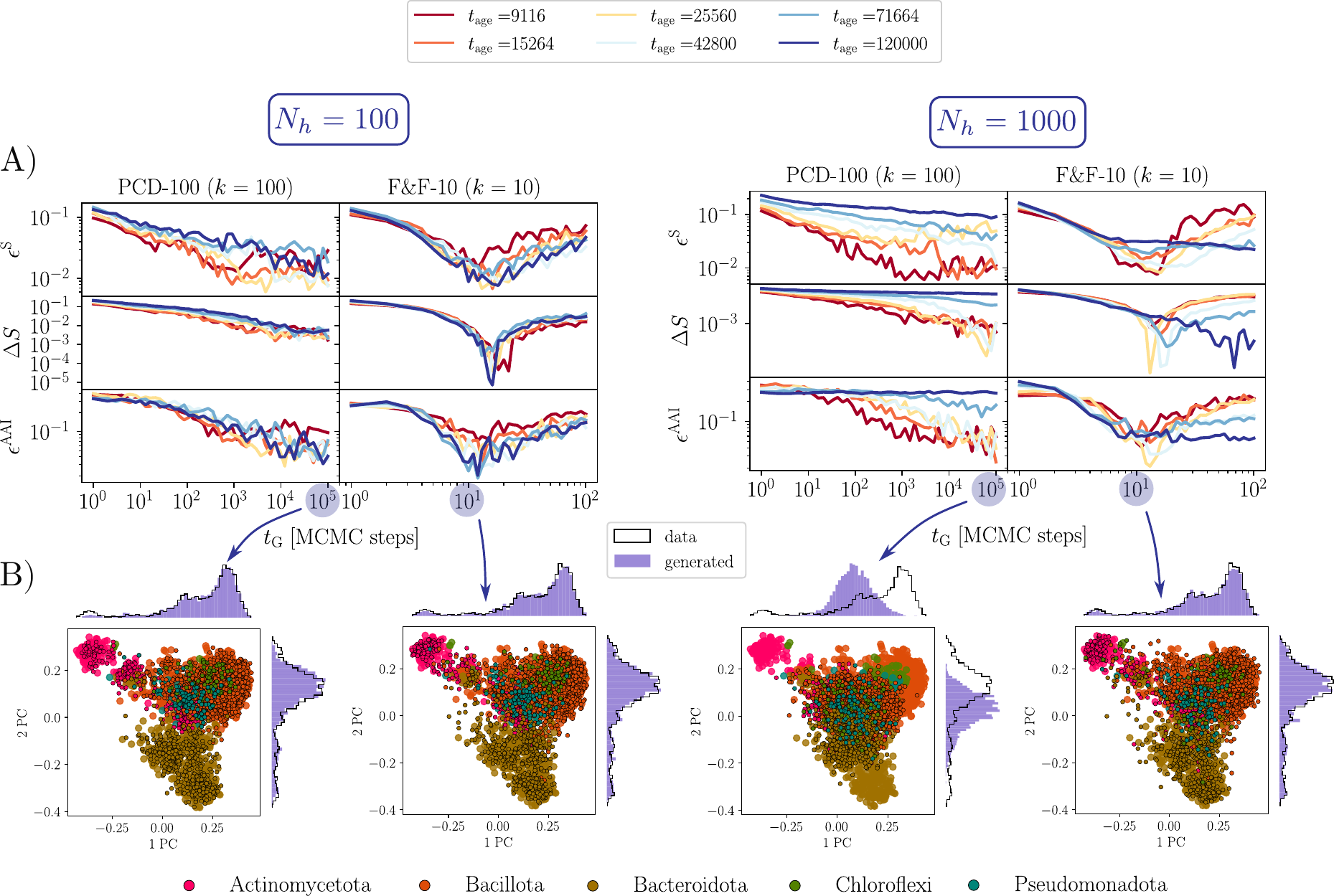}
    \caption{Example of instability of the PCD RBM on the SAM dataset when the number of hidden nodes, $N_h$, is changed. In A) we show the comparison between the score curves of PCD and F\&F RBMs for $N_h=100$ (left) and $N_h = 1000$ (right). In B), instead, we show the data generated with the most trained model (dark-blue curve) after $10^5$ and 10 MCMC updates for PCD and F\&F, respectively. The data are projected along the two principal components of the dataset's PCA. The big colored dots represent the true data, whereas the contoured dots represent the generated samples. Different colors correspond to different categories. The histograms on the sides show the distributions of the true data (black profile) and the synthetic data (violet-shaded area) projected along the two principal directions of the dataset.  For $N_h=100$ we obtained a properly trained model using both the training methods, and these models are those shown in the main text. Conversely, when we set $N_h=1000$ we obtained a PCD RBM model displaying huge thermalization times after just a few training epochs, while F\&F still managed to yield a model whose generated data correctly span the whole data space.}
    \label{fig:scores comparison hidden nodes}
\end{figure*}

\begin{figure*}[!t]
    \centering
    \includegraphics[width=0.8\linewidth]{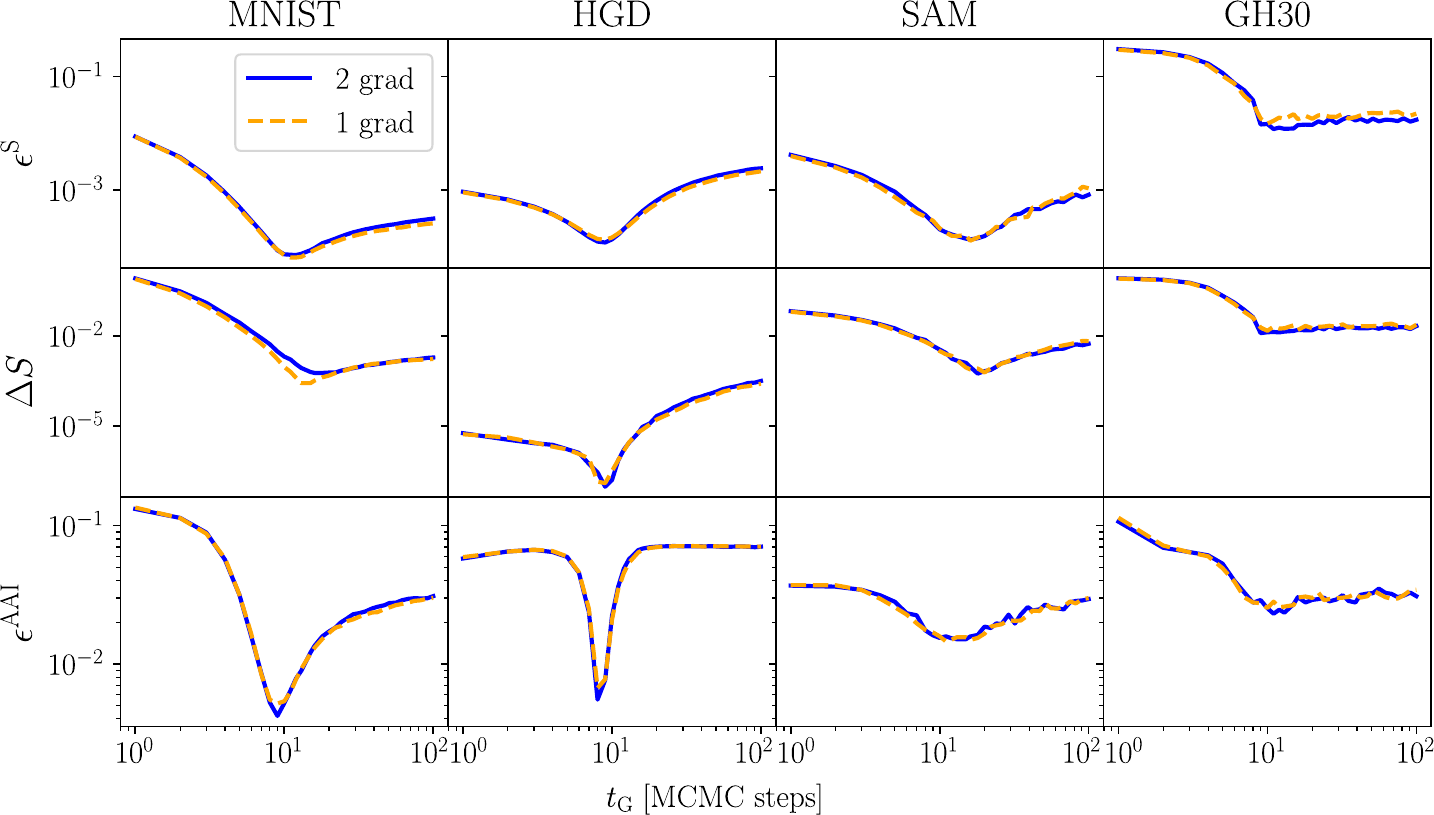}
    \caption{Comparison of the generation capabilities between a F\&F model trained with both the gradients (blue solid line) and one that only uses the gradient computed by clamping the labels to the data (dashed orange line).}
    \label{fig:1grad vs 2grad - generation}
\end{figure*}

\begin{figure*}[!t]
    \centering
    \includegraphics[width=0.8\linewidth]{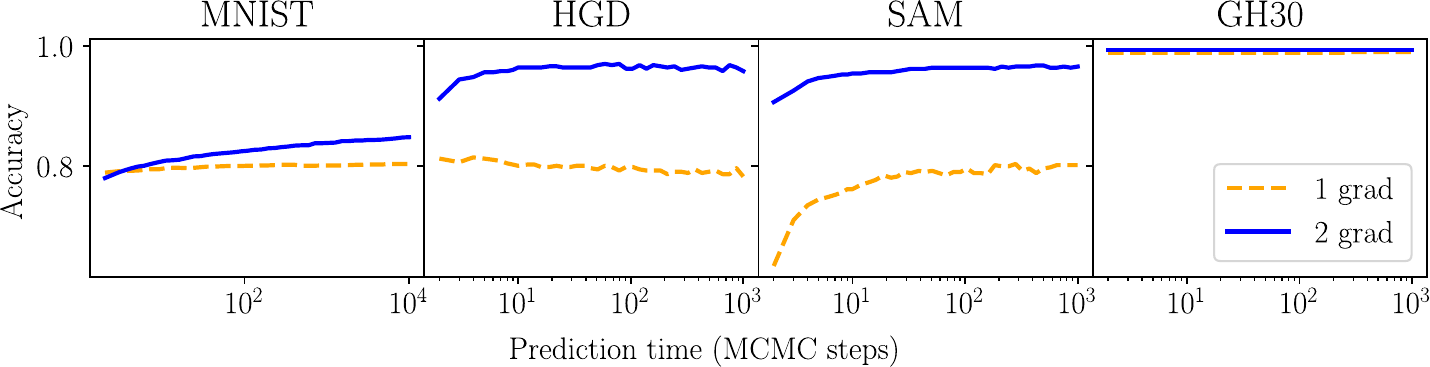}
    \caption{Comparison of the prediction capabilities between a F\&F model trained with both the gradients (blue solid line) and one that only uses the gradient computed by clamping the labels to the data (dashed orange line).}
    \label{fig:1grad vs 2grad - prediction}
\end{figure*}

\begin{figure*}[!t]
    \centering
    \includegraphics[width=0.8\linewidth]{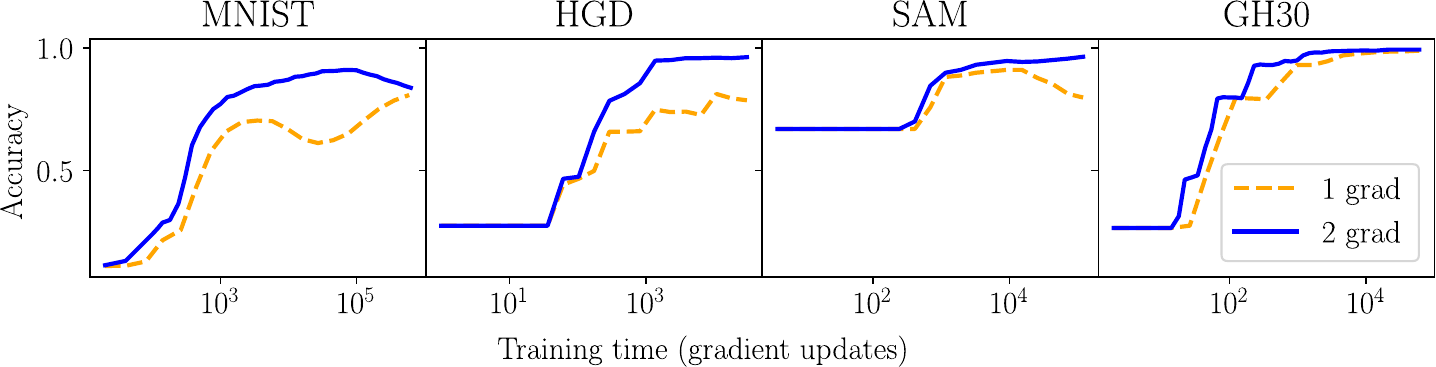}
    \caption{Comparison of the prediction capabilities between a F\&F model trained with both the gradients (blue solid line) and one that only uses the gradient computed by clamping the labels to the data (dashed orange line) as a function of the training time. For each machine, the prediction time is set to 1000 Monte Carlo steps.}
    \label{fig:1grad vs 2grad - training}
\end{figure*}

\end{document}